\relax
\documentclass[letterpaper]{article} 
\usepackage{aaai20}  
\usepackage{times}  
\usepackage{helvet} 
\usepackage{courier}  
\usepackage[hyphens]{url}  
\usepackage{graphicx} 
\usepackage{graphicx}  
\usepackage{amsmath} 

\title{A Density-Aware PointRCNN for 3D Object Detection in Point Clouds}
\author{Jie Li\textsuperscript{\rm 1,2}, Yu Hu\textsuperscript{\rm 1,}\textsuperscript{\rm 2}\thanks{Corresponding author: Yu Hu, huyu@ict.ac.cn. This work is supported in part by the National Key RD Program of China under grant No. 2018AAA0102701, in part by the Science and Technology on Space Intelligent Control Laboratory under grant No. HTKJ2019KL502003, and in part by the Innovation Project of Institute of Computing Technology, Chinese Academy of Sciences under grant No. 20186090.}\\
\textsuperscript{\rm 1}Research Center for Intelligent Computing Systems \\
Institute of Computing Technology, Chinese Academy of Sciences\\ 
\textsuperscript{\rm 2}University of Chinese Academy of Sciences\\
lijie2019@ict.ac.cn, huyu@ict.ac.cn 
}
 
\begin{document}

\maketitle

\begin{abstract}
We present an improved version of PointRCNN for 3D object detection, in which a multi-branch backbone network is adopted to handle the non-uniform density of point clouds. An uncertainty-based sampling policy is proposed to deal with the distribution differences of different point clouds. The new model can achieve about 0.8 AP higher performance than the baseline PointRCNN on KITTI val set. In addition, a simplified model using a single scale grouping for each set-abstraction layer can achieve competitive performance with less computational cost.
\end{abstract}

\section{Introduction}
3D object detection is one of the crucial techniques in autonomous driving. The relatively mature 2D object detection generally gives an axially aligned 2D bounding box on the image plane, but 3D object detection can give an oriented 3D bounding box in the 3D space. The accurate spatial location information allows autonomous vehicles to effectively predict and plan behaviors and paths to avoid collisions and violations.

\begin{figure}[t]
\centering
\includegraphics[width=0.99\columnwidth]{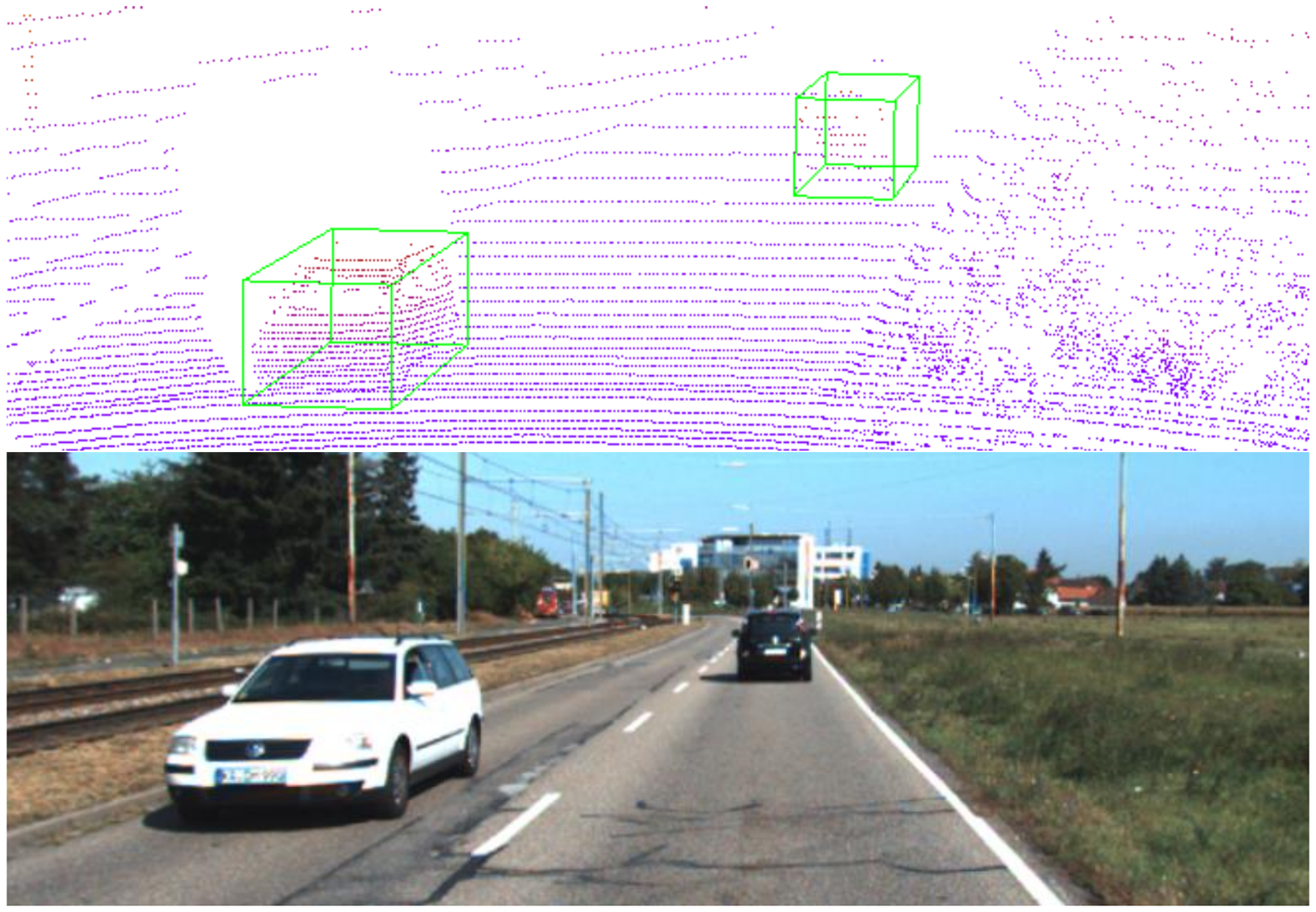} 
\caption{Illustration of the non-uniform density of point clouds, with the point cloud on the top and the corresponding image on the bottom. Two 3D bounding boxes are plotted in the point cloud for the two cars shown in the image. The near car has much denser points than the far one.}
\label{fig1}
\end{figure}

Compared with other sensors, such as monocular or depth cameras, LiDAR can give more accurate range information and can be used in outdoor scenes. The working principle of LiDAR determines the inherent characteristics of the point cloud, such as sparsity, irregularity, large amount of data, and non-uniform density, and we try to tackle the non-uniformity in this paper, as is illustrated in Fig.~\ref{fig1}.

In addition to point clouds, some 3D detection methods also use RGB images from cameras. For example, MV3D~\cite{chen2017multi} and AVOD~\cite{ku2018joint} encode 3D information into multiple views (LiDAR bird view and front view, camera images, \emph{etc.}). PC-CNN~\cite{du2018a}, PointFusion~\cite{xu2018pointfusion}, F-PointNet~\cite{qi2018frustum}, Frustum ConvNet~\cite{wang2019frustum}, Roarnet~\cite{shin2019roarnet} rely on the proposals from the 2D detector for RGB images, and only points in proposals are delivered to subsequent stages. IPOD~\cite{yang2018ipod} requires the semantic map from 2D segmentation network to score anchor boxes. ContFuse~\cite{liang2018deep}, MMF~\cite{liang2019multi} consider pixel-level feature fusion of point cloud bird view and RGB images. The above detectors need both point clouds and RGB images, and their performance heavily relies on the 2D detection performance. Recently, more and more newly proposed methods prefer to use only point cloud as input, such as PointRCNN~\cite{shi2019pointrcnn}, Part-$A^{2}$~\cite{shi2019part}, STD~\cite{yang2019std}, PV-RCNN~\cite{shi2020pv}, 3DSSD~\cite{yang20203dssd}, SASSD~\cite{he2020structure}, and we choose one of them (PointRCNN) to study how to deal with the non-uniform density of point clouds.

The leading 3D detection methods can generally be divided into two categories, the grid-based methods and the point-based methods. The grid-based methods transform the irregular sparse 3D point cloud into a regular compact representation, such as 3D voxels or 2D bird-eye-view (BEV) images, which employ 3D or 2D CNN to learn features. The features of BEV images are often defined manually, like occupancy, height or reflectance, which are used in MV3D, AVOD, PIXOR~\cite{yang2018pixor}, YOLO3D~\cite{ali2018yolo3d}, Complex-YOLO~\cite{simon2018complex}, LaserNet~\cite{meyer2019lasernet}. The features of voxels are also defined manually in early methods, such as Vote3Deep~\cite{engelcke2017vote3deep} and 3DFCN~\cite{li20173d}, but after the propose of Voxel Feature Encoding (VFE) in VoxelNet~\cite{zhou2018voxelnet}, later methods usually use VFE to learn features, such as SECOND~\cite{yan2018second}, PointPillars~\cite{lang2019pointpillars}, Fast Point R-CNN~\cite{chen2019fast}, Patch Refinement~\cite{lehner2019patch}, Voxel-FPN~\cite{kuang2020voxel}, Part-$A^{2}$. On the other hand, the point-based methods can process the raw point cloud data directly, in which PointNet~\cite{charles2017pointnet} or PointNet++~\cite{qi2017pointnet} are often used to extract the features of points, such as PointFusion, F-PointNet, Frustum ConvNet, Roarnet, IPOD, PointRCNN, STD, 3DSSD. The irregularity, sparsity, voluminousness of point clouds are relatively well considered in those methods, but the non-uniform density has not been emphasized enough, which is more obvious in point-based methods, and this inspires us to study the treatment of non-uniformity of points for point-based methods.

\begin{figure*}[t]
\centering
\includegraphics[width=0.99\textwidth]{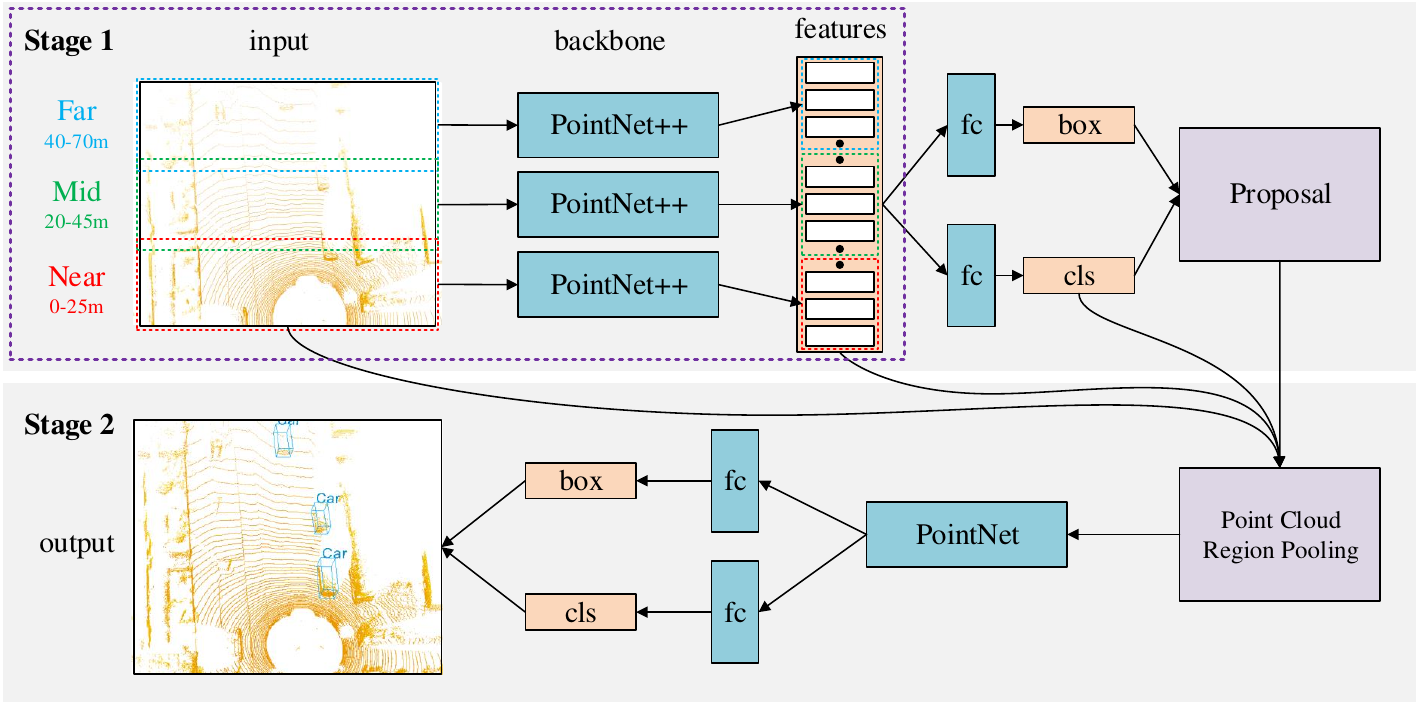} 
\caption{The overall architecture of the proposed Density-Aware PointRCNN, with the different part from the PointRCNN enclosed in a purple dashed rectangle. The detector consists of two stages: stage 1 for generating 3D proposals from point clouds, and stage 2 for refining the proposals. In stage 1, a three-branch backbone network is adopted, and accordingly, the input point cloud is divided into three parts (0-25m, 20-45m, 40-70m), with 5m overlap of adjacent regions.}
\label{fig2}
\end{figure*}

In point clouds, the car in near areas has much more points than that in far areas, so we designed a multi-branch backbone network, which can extract the features of near, mid, and far points separately. In addition, the difference in point distributions of various point clouds can be large, leading to significant changes in the number of points within a certain distance interval. The number change is a problem for each branch, which need a fixed number of input points. In order to balance the sampling effectiveness and diversity, we propose an uncertainty-based sampling policy to deal with the problem. With the multi-branch backbone adapting to various density, a simplified model using only a single scale grouping can be implemented, which can achieve similar performance with less computation.

Our contributions can be summarized into three-fold:
\begin{itemize}
\item A three-branch backbone network is designed to extract features of points in near, mid and far areas separately.
\item An uncertainty-based sampling policy is put forward for dealing with the number change of points in different point clouds.
\item A simplified detector with only a single scale grouping for each set-abstraction layer is implemented, showing similar performance but less computation.
\end{itemize}

\section{Related Work}
Although there have been many researches on the 3D detection methods from point clouds, the density non-uniformity of points has not been paid enough attention. Only a small amount of previous work has implicitly considered this issue:

1) PointNet++~\shortcite{qi2017pointnet} proposes a multi-scale grouping (MSG) strategy to construct density adaptive PointNet layers. This strategy extracts multiple scales of local patterns at each abstraction level and combine them to enhance the robustness of feature learning under non-uniform sampling density. However, it does not explicitly consider the influence of distance on the density (the same parameters are used in different regions), and the multi-scale grouping increases the computational complexity.

2) Frustum ConvNet~\shortcite{wang2019frustum} extracts features at different distances by a sequence of frustums for each region proposal, and the frustums are used to group local points. This method has considered the distance, but it rely on 2D proposals in RGB images, and the large grouping granularity is not conducive to local feature extraction.

3) Voxel-FPN~\shortcite{kuang2020voxel} performs multi-resolution voxelization on the original point cloud, and then a FPN~\cite{lin2017feature} structure is adopted to fuse multi-resolution features, which is similar to the multi-scale grouping in PointNet++ and the factor of distance is not considered.

4) RT3D~\cite{zeng2018rt3d} considers various amounts of valid points in different parts of the car. For example, the LiDAR data will concentrate on the right side of the car if it is in the left front of the LiDAR, but the density non-uniformity of the whole point cloud is not considered in this method.

Recently, some studies began to focus on the density non-uniformity of point clouds:

1) RangeAdaption~\cite{wang2019range} explores cross-range adaption for 3D object detection using LiDAR, which uses an adversarial global adaptation and a fine-grained local adaptation to make the features of far-range objects similar to that of near-range objects. This method can improve the performance on the far-range objects without adding auxiliary parameters, but it ignores the uniqueness of far object itself.

2) DistanceDependent~\cite{engels20203d} directly trains two separate detectors to extract features of close-range and long-range objects, which leads to improvements for objects in 0-35 meter range and 35-70 meter range. However, training two separate networks splits the correlation between point clouds of different densities and affects the generalization performance of the model.

3) SegVoxelNet~\cite{yi2020segvoxelnet} designs a depth-aware head with convolution layers of different kernel sizes and dilated rates, to explicitly model the distribution differences, which includes three part and each part is made to focus on its own object detection range. The backbone network is shared for objects at different ranges, and the depth-aware feature extraction only used in head network, which limits the improvement of performance.

As an important characteristic of point cloud distribution, density non-uniformity is a key factor that affects data distribution and thus limits the performance of detection networks. At present, there is still a lack of in-depth research for non-uniformity of point clouds, and this paper presents a feasible approach to deal with the issue.

\section{A Density-Aware PointRCNN}
In this section, we present the proposed Density-Aware PointRCNN, which is abbreviated as DA-PointRCNN. The overall architecture is illustrated in Fig.~\ref{fig2}.

As shown in Fig.~\ref{fig2}, the DA-PointRCNN has the same basic framework as that of PointRCNN, which consists of two stages: stage 1 sub-network for generating 3D proposals from raw point cloud, and stage 2 sub-network for refining the proposals. The different part of DA-PointRCNN from PointRCNN is enclosed in the purple dashed rectangle, in which a three-branch backbone network is adopted, and the input point cloud is divided into three parts (0-25m, 20-45m, 40-70m), with 5m overlap of adjacent regions. The point features extracted from three regions by different branches are then concatenated to be the input features of RPN.

\begin{figure}[t]
\centering
\includegraphics[width=0.99\columnwidth]{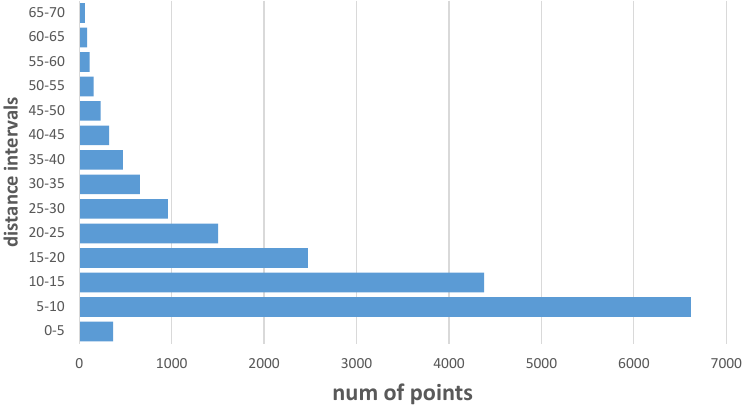} 
\caption{The average number of points per point cloud at each range on the \emph{train} split of KITTI dataset.}
\label{fig3}
\end{figure}

\subsection{A Three-Branch Backbone Network}
In SSD~\cite{liu2016ssd} for 2D detection, the features of objects at different scales come from different feature maps. In point clouds, the sizes of vehicles are invariant at different ranges, but the density of points is quite different as shown in Figure. Accordingly, we can extract separate features for objects with different point densities by a multi-branch backbone network.

The KITTI dataset~\cite{geiger2012are} contains 7,481 training samples and 7,518 test samples. The data settings in this paper are the same as PointRCNN~\shortcite{shi2019pointrcnn}, in which the training samples are split into \emph{train} split (3,712 samples) and \emph{val} split (3,769 samples). We counted the number of points in \emph{train} split, and the average number of points per point cloud at each range is shown in Fig.~\ref{fig3}.

Considering distribution similarity and division balance, the point cloud is divided into three ranges: 0-20m, 20-40m, and 40-70m, corresponding to near, mid, and far regions. In order to learn explicitly the distribution differences of points at different ranges, a three-branch backbone network is designed to extract features of points in near, mid and far regions separately.

For training phase, a 5m overlap area (slightly larger than the average size of cars) between adjacent regions is introduced to avoid objects being cut off at the boundary, thus the adjusted ranges are: 0-25m, 20-45m, and 40-70m. For inference phase, the overlap area shrinks to 3m to get more focused branches and better performance, but sometimes we'll keep the results with 5m overlap if its performance is occasionally better.

\subsection{An Uncertainty-based Sampling Policy}

Following PointRCNN~\shortcite{shi2019pointrcnn}, 16,384 points are sampled from each point-cloud scene as the inputs, and points are randomly repeated if the number of points is fewer than 16,384.

With a three-branch backbone network, the input points are required to be divided into three parts. A natural division strategy is based on the point density, \emph{i.e.}, the number of sampling points is proportional to the number of points in each region. For \emph{train} split, we count the mean value of points in each region, and get $m_1$, $m_2$, $m_3$ for near, mid, and far regions respectively ($m_1=13.8k$, $m_2=3.6k$, $m_3=1.0k$). Because the farther point is more important, the 16,384 points can be divided into 11,264 (near), 4,096 (mid), and 1,024 (far) accordingly. However, the natural division strategy results in poor performance, especially for mid and far objects.

With a further analysis on the distribution of points, we found that, for each region, the number of points in different point clouds may vary greatly. To measure the uncertainty, the standard deviations of number of points in different regions are calculated, which are $\sigma_1$, $\sigma_2$, and $\sigma_3$ for near, mid and far regions ($\sigma_1=1.8k, \sigma_2=1.1k, \sigma_3=0.5k$). We can see that the change ratio of points in far regions is about 50\% (\emph{i.e.} $\sigma_3/m_3$), affecting greatly the performance of the natural division strategy.

Given standard deviations, an uncertainty-based sampling policy is put forward to deal with the number change of points in different point clouds, in which the 16,384 points are divided into 9,216 (near), 5,120 (mid), and 2,048 (far). Actually, this division is from the following strategy 4, in which the near, mid, and far regions are balanced well:
\begin{itemize}
\item strategy 1: $m_2+\sigma_2$ for mid, $m_3+\sigma_3$ for far regions
\item strategy 2: $m_2+1.5\sigma_2$ for mid, $m_3+1.5\sigma_3$ for far regions
\item strategy 3: $m_2+2\sigma_2$ for mid, $m_3+2\sigma_3$ for far regions
\item strategy 4: $m_2+1.5\sigma_2$ for mid, $m_3+2\sigma_3$ for far regions
\end{itemize}

The above four division strategies are compared in the Experiments section.

\subsubsection{Other Improvement Tricks}

To cooperate with the three-branch backbone network, four tricks are adopted: adjusted radii, shared RPN, extra training, and joint training.

\begin{itemize}
\item Adjusted Radii: the grouping radius setting for the first set-abstraction layer of PointRCNN is $(0.1, 0.5)$, and in the new model, the radius settings for first set-abstraction layer are $(0.1, 0.5), (0.2, 0.6), (0.4, 0.8)$ for near, mid, and far branches. The following set-abstraction layers are adjusted accordingly. Though this adjustment brings no obvious accuracy improvement, it makes the model performance more stable in experiments.
\item Shared RPN: each branch of the backbone network focuses on learning features of points in its region, so a shared RPN for three branches helps to provide unified feature representation for RCNN, improving the generalization performance of the model.
\item Extra Training: the three-branch network has about 3 times the parameters of the original one, so the bigger backbone network should require more training epochs to converge. In experiments, we just doubled the training epochs.
\item Joint Training: the backbone network and RPN trained in the first stage, will also be trained jointly with the RCNN in the second stage, which is shown to be effective in experiments.
\end{itemize}

\subsection{A Simplified Detector with a Single Scale Grouping}
As is stated in Related Work, the multi-scale grouping (MSG) strategy proposed in PointNet++ is to deal with the non-uniform sampling density. The MSG strategy approach is computationally expensive since it need to extract features of different scales for each centroid point, \emph{e.g.} 2 scales for PointRCNN~\shortcite{shi2019pointrcnn} and 3 scales for STD~\shortcite{yang2019std}.

In this paper, the multi-branch backbone network is designed for the same purpose of tackling non-uniformity, so the role of MSG strategy is not as important as that in the original PointRCNN. Therefore, a simplified detector with only a single scale grouping for each set-abstraction layer is proposed in this section, showing similar performance but requiring less computation, which demonstrates the effectiveness of multi-branch networks for the treatment of non-uniformity in point clouds. For convenience, the simplified version of DA-PointRCNN is abbreviated as sDA-PointRCNN.

The improvement tricks stated in the previous section are also used in the simplified detector, \emph{i.e.} the adjusted radii, shared RPN, extra training, and joint training. Specifically, the radius settings for the 4 set-abstraction layers of PointRCNN is $(0.1, 0.5)$, $(0.5, 1.0)$, $(1.0, 2.0)$, $(2.0, 4.0)$, and the settings for the three-branch backbone network of the simplified detector are as below:
\begin{itemize}
\item branch for near regions: $(0.4), (0.8), (1.6), (3.2)$
\item branch for mid regions: $(0.8), (1.6), (3.2), (4.0)$
\item branch for far regions: $(1.0), (2.0), (3.0), (4.0)$
\end{itemize}

\subsection{Loss Function}

For fair comparison, the loss function of PointRCNN is also adopted in DA-PointRCNN and its simplified version.

For stage 1 sub-network, focal loss~\cite{lin2017focal} is used to address the foreground-background class imbalance as Eq.~\ref{eq:focal}, and the full bin-based loss~\shortcite{shi2019pointrcnn} is used for generating 3D box proposals as Eq.~\ref{eq:bin}.

\begin{equation}
{L_{{\text{focal}}}}\left( {{p_t}} \right) =  - {\alpha _t}{\left( {1 - {p_t}} \right)^\gamma }\log \left( {{p_t}} \right)
\label{eq:focal}
\end{equation}
where $p_t$ is the probability of correct classification.

\begin{equation}
{L_{{\text{reg}}}} = \frac{1}{{|P|}}\sum\nolimits_{p \in P} {\left( {L_{{\text{bin}}}^{\left( p \right)} + L_{{\text{res}}}^{\left( p \right)}} \right)}
\label{eq:bin}
\end{equation}
where $P$ is the set of foreground points, and $|P|$ is the number of points in set.

For stage 2 sub-network, cross entropy loss is used to for classification and full bin-based loss is adopted for regression, as Eq.~\ref{eq:refine}.
\begin{equation}
{L_{{\text{refine}}}} = \frac{1}{{|O|}}\sum\nolimits_{o \in O} {{F_{cls}}\left( {{p_o},{l_o}} \right)}  + \frac{1}{{|R|}}\sum\nolimits_{r \in R} {\left( {\tilde L_{{\text{bin}}}^{\left( r \right)} + \tilde L_{{\text{res}}}^{\left( r \right)}} \right)}
\label{eq:refine}
\end{equation}
where $O$ is the set of 3D proposals, and $R$ is the set of positive proposals.

\section{Experiments}
In this section, the DA-PointRCNN is evaluated on the widely used KITTI Object Detection Benchmark~\shortcite{geiger2012are}. The dataset contains three main classes, namely, cars, pedestrians and cyclists, and only the car class is considered in this paper to demonstrate the performance of the proposed method. Firstly, the implementation details of DA-PointRCNN is introduced. Then the main results on KITTI \emph{val} split are shown. Finally, ablation studies are conducted to analyze the effectiveness of components in the new model.
\begin{table*}[t]
\caption{Performance comparison of 3D object detection on the car class of KITTI val split set. The evaluation metric is Average Precision(AP) with IoU threshold 0.7. The PointRCNN with * is the one in the original paper, and the PointRCNN without * is the re-implemented version in our project.}\smallskip
\centering
\smallskip\begin{tabular}{c||c c c |c}
\hline
Method & & Car (IoU=0.7) &  &Backbone + RPN  \\
 & Easy & Moderate & Hard & (ms) \\
\hline
*PointRCNN & 88.88 & 78.63 &  77.38 & -\\
PointRCNN & 87.79 & 77.62 & 76.68 & 43 \\
DA-PointRCNN & 88.21 & \textbf{78.41} & \textbf{77.20} & 58\\
sDA-PointRCNN & \textbf{88.26} & 78.13 & 76.66 & 42\\
\hline
\end{tabular}
\label{table1}
\end{table*}


\subsection{Implementation Details}
The proposed model is modified from the open source PointRCNN https://github.com/sshaoshuai/PointRCNN. To facilitate comparative analysis, the main experimental setup remains the same as the original model, only the parts involving the new model are changed accordingly.

\subsubsection{Network Architecture}
The stage-2 sub-network (RCNN) remains unchanged, only the backbone for the stage-1 sub-network (RPN) is changed from one branch to three branches. Each branch shares the same configures with others except for the number of points and the grouping radius. The input number of points of each branch is determined by the uncertainty-based sampling policy. The total 16,384 points are divided into 9,216 (near), 5,120 (mid), and 2,048 (far). Correspondingly, the four set-abstraction layers sample points into groups with sizes 2,304-576-144-36 for near branch, 1,280-320-80-20 for mid branch, and 512-128-32-8 for far branch.

In addition, the proposal ratios of different regions are adjusted into 0.3 (near), 0.5 (mid), and 0.2 (far), which is based on the average number of objects in each region in the \emph{train} split.

\subsubsection{The Training Scheme}
Same as PointRCNN, the bin-based proposal generation and refinement are adopted. The data augmentation of random flip, scaling, rotation, and the GT-AUG which adds extra non-overlapping ground-truth boxes from other scenes are also employed.

The two stage sub-networks are also trained separately, but the training strategy is the strategy (a) presented in the project PointRCNN. There are two strategies to train stage-2 sub-network (RCNN), strategy (a) using online GT augmentation and strategy (b) using offline GT augmentation. The best model provided by the project is trained by the offline augmentation strategy, in which more resources are used to save and shuffle the RoIs and features to train the RCNN network. The performance is not stable due to the small training/val set.

The strategy (a) adopted in this paper for the baseline model is more elegant and easy to train, and the performance is about 1.2\% AP lower than the one provided by the project. Our new model is also trained by strategy (a), which is convenient to show the improvement of our method. The improvement trick of joint training is based on strategy (a), and for the second stage, the RCNN and RPN are trained jointly.

\subsection{Main Results on KITTI}
The 3D detection dataset of KITTI contains 7,481 training samples and 7,518 test samples. Same as PointRCNN, the training samples are split into \emph{train} split (3,712 samples) and \emph{val} split (3,769 samples). For 3D object detection, we compare the proposed method (DA-PointRCNN and sDA-PointRCNN) with PointRCNN on the \emph{val} split of KITTI dataset, and the results are shown in Tab.~\ref{table1}. The PointRCNN model evaluated in this section is trained from the open source project with strategy (a), and its performance is about 1.2\% AP lower than that in paper.

The DA-PointRCNN can achieve about 0.8 AP higher than the baseline PointRCNN at moderate difficulty, about 0.5 AP higher for the other two difficulties. The sDA-PointRCNN shows similar performance to PointRCNN at hard difficulty, and better performance (about 0.5 AP higher) for easy and moderate difficulties.

Comparing with DA-PointRCNN, its simplified version, \emph{i.e.} sDA-PointRCNN, can achieve similar performance for easy situation, but slightly worse for moderate and worse for hard situations. The performance differences should result from the distribution differences of point clouds. Comparing with the easy situation, the non-uniformity of the other two situations is more serious, in which the MSG (multi-scale grouping) plays a more important role.

As for the running time, the stage-1 subnetworks (backbone+RPN) of three models are compared on a single GeForce RTX 2080Ti GPU. According to the network structure, the running time of DA-PointRCNN should be similar to that of PointRCNN, and sDA-PointRCNN should be faster. However, as shown in Tab.~\ref{table1}, the subnetwork of PointRCNN has similar running time to that of sDA-PointRCNN, and the DA-PointRCNN is slower. The cause of the phenomenon is the implementation of multi-branch backbone, currently the features of three branches are calculated serially rather than in parallel, which can be sped up by parallel implementation.

\subsection{Ablation Study}
In this section, ablation experiments are carried out to analyze the proposed model. Firstly, the settings of of sampling policy for the multi-branch backbone network are compared and discussed. Then the effectiveness of different components in DA-PointRCNN is demonstrated. All experiments are trained on the \emph{train} split and evaluated on the \emph{val} split with the car class.

\subsubsection{The Uncertainty-based Sampling Policy.}
As is stated in previous section, four uncertainty-based sampling policies are designed to deal with the number change of points in different point clouds. The results are shown in Tab.~\ref{table2}, and the original strategy is the one without considering the uncertainty, in which $m_2$ for mid and $m_3$ for far regions.

Compared with other strategies, strategy 4 can provide a better balance among near, mid and far regions, which is adopted in the proposed DA-PointRCNN and sDA-PointRCNN. The sampling strategy pays more attention to mid and far points, but not aggressive as strategy 2, which leaves less points for near regions.

\begin{table}[t]
\caption{Performance for different sampling strategies. $AP_E, AP_M, AP_H$ denote the Average Precision(AP) with IoU threshold 0.7 for easy, moderate, hard difficulty on \emph{val} split.}\smallskip
\centering
\smallskip\begin{tabular}{c||c c c}
\hline
Sampling Strategy &  $AP_E$ & $AP_M$ &  $AP_H$    \\
\hline
original   & 87.44 & 77.27 & 75.04 \\
strategy 1 & 87.14 & 77.47 & 75.77 \\
strategy 2 & 88.02 & 77.86 & 76.01 \\
strategy 3 & 87.14 & 77.72 & 75.36 \\
strategy 4 & \textbf{88.08} & \textbf{78.13} & \textbf{76.65} \\
\hline
\end{tabular}
\label{table2}
\end{table}

\subsubsection{The Improvement Tricks}
The effectiveness of three improvement tricks are analyzed in this part: shared RPN, extra training, and joint training. The $AP_M$, as the most stable index of three (\emph{i.e.} $AP_E, AP_M, AP_H$), is used to measure the performance improvement, as is shown in Tab.~\ref{table3}.

As is mentioned previously, each branch of the backbone network focuses on learning features of points in its region, and the shared RPN helps to provide unified feature representation, which is beneficial to the model generalization ability.

Extra training is used to train a larger backbone network, but the improvement of using it alone is limited. With both extra training and joint training adopted, the model can obtain a better performance. The joint training can provide a better balance between stage-1 and stage-2 sub-networks, preventing network from falling into local optimum prematurely due to greedy search.

\begin{table}[t]
\caption{Performance for different improvement tricks. $AP_M$ denote the Average Precision(AP) for moderate difficulty.}\smallskip
\centering
\smallskip\begin{tabular}{c|c|c|c c c}
\hline
Shared & Extra & Joint &  $AP_M$    \\
RPN & Training & Training & \\
\hline
$\times$     & $\times$      & $\times$     & 78.13  \\
$\surd$ & $\times$      & $\times$     &  78.20  \\
$\surd$ & $\surd$  & $\times$     & 78.23  \\
$\surd$ & $\surd$  & $\surd$ &  \textbf{78.41} \\
\hline
\end{tabular}
\label{table3}
\end{table}


\section{Conclusion}
We have presented DA-PointRCNN, an improved version of PointRCNN for 3D object detection from point clouds. The new model adopts a three-branch backbone network to handle the non-uniform density of point clouds. To cooperate with the backbone, we propose an uncertainty-based sampling policy to deal with the distribution differences of different point clouds. Experiments and ablation studies demonstrate the effective of the new model, which can achieve about 0.8 AP higher performance than the baseline PointRCNN on KITTI val set. In addition, a simplified model using a single scale grouping for each set-abstraction layer is designed, which can achieve similar performance with less computational cost.

\bibliography{Bibliography-File}
\bibliographystyle{aaai}

\end{document}